\documentclass[10pt, a4paper]{article}
\usepackage{xspace} 
\usepackage{multirow}
\usepackage{tabularx}
\usepackage{CJK}

\usepackage[final]{lrec-coling2024} % this is the new style
\usepackage{amsmath}
\title{Korean Bio-Medical Corpus (KBMC) for Medical Named Entity Recognition}
\usepackage{makecell}

\name{\begin{tabular}{c}
      Sungjoo Byun\textsuperscript{1}, Jiseung Hong\textsuperscript{2}, Sumin Park\textsuperscript{1}, Dongjun Jang\textsuperscript{1},\\
      Jean Seo\textsuperscript{1}, Minseok Kim\textsuperscript{1}, Chaeyoung Oh\textsuperscript{1}, Hyopil Shin\textsuperscript{1}
      \end{tabular}}

\address{\textsuperscript{1}Seoul National University \\
         \{byunsj, mam3b, qwer4107, seemdog, snumin44, nyong10, hpshin\}@snu.ac.kr\\
         \textsuperscript{2}KAIST \\
         jiseung.hong@kaist.ac.kr}

\abstract{
Named Entity Recognition (NER) plays a pivotal role in medical Natural Language Processing (NLP). Yet, there has not been an open-source medical NER dataset specifically for the Korean language. To address this, we utilized ChatGPT to assist in constructing the KBMC (Korean Bio-Medical Corpus), which we are now presenting to the public. With the KBMC dataset, we noticed an impressive 20\% increase in medical NER performance compared to models trained on general Korean NER datasets. This research underscores the significant benefits and importance of using specialized tools and datasets, like ChatGPT, to enhance language processing in specialized fields such as healthcare.
 \\ \newline  }

\begin{document}

\maketitleabstract

\section{Introduction}

The significance of domain-specific Named Entity Recognition (NER), especially in fields like law and medicine, calls for more in-depth research and investigation. The role of NER in medical NLP is as follows: Firstly, NER contributes to processing medical terminology. Medical NER enables language models to identify and process medical terminologies and jargon. Next, it facilitates information extraction from unstructured data. In fact, \citet{unstructured} have performed NER to remove or encode information from an unstructured medical dataset. Moreover, NER contributes to entity identification and the anonymization of sensitive patient-specific information \citep{italian}. 

However, it is problematic that medical NER datasets are insufficient. This problem becomes even more challenging as domain-specific NER tasks require extensive labeling, particularly for specific entity categories like Disease, Body, and Treatment. The difficulty is further amplified due to the necessity of expert-level knowledge in medical domains. The data scarcity issue worsens in relatively low-resource languages like Korean. The fact that there is no open-source medical NER dataset for Korean demonstrates the severity of the problem. In order to resolve the data scarcity problem, we introduce KBMC (Korean Bio-Medical Corpus), the first open-source medical NER dataset for Korean. We utilize ChatGPT\footnote{\url{https://chat.openai.com}} for effective sentence creation. Subsequently, we annotate entities corresponding to disease name, body part, and treatment following the BIO format. To augment the dataset and to check the performance in general text as well, we concatenate the Naver dataset,\footnote{\url{https://github.com/naver/nlp-challenge}} which is the Korean NER dataset with our KBMC in the experiment. 

In our research, we evaluate the effectiveness and utility of KBMC by comparing the performance of multiple language models. These models either use a general NER dataset (solely the Naver NER dataset) or a domain-specific dataset (a combination of the Naver NER dataset and KBMC). The results demonstrate that our dataset significantly enhances the accurate recognition of medical entities by more than 20 percent.

Contributions of our research are as follows: 
\begin{itemize}
\setlength\itemsep{0 em}
\item We describe and publicly release Korean Bio-Medical Named Entity Recognition Corpus (KBMC), the first open-source Korean medical NER dataset. This contributes to solving the data scarcity problem.
\item Our research aims to play crucial role in medical data processing. Medical NER would facilitate the sensitive data anonymization process and contribute to the reconstruction of medical data that lack standardized formats.
\end{itemize}

\section{Related Work}
\subsubsection*{Medical NER}
As a part of the entity representation task, various studies, mainly in English, have explored the medical field. Traditional research has conducted bio-medical NER using Long short-term memory (LSTM) models (\citealp{rnn}; \citealp{lstm_ner}; \citealp{lstm2}). \citet{elmo} test Biomedical Language Understanding Evaluation (BLUE) benchmark, including NER with BERT and ELMo. Bio-BERT, a pre-trained language representation model for biomedical text mining, shows high performance in bio-NER \citep{bio_bert}. 
Also, various toolkits that facilitate clinical NER implemented using SpaCy \citep{medspacy}, Apache Spark \citep{spark}, and Flair \citep{flair} have been introduced.

\subsubsection*{Medical NER dataset} 
Medical NER is crucial, as shown by numerous medical concept extraction challenges, such as those hosted by i2b2 \citep{NER_i2b2} and n2c2 \citep{ner_n2c2}. To tackle the data scarcity in the medical field, SemClinBER, a Portuguese medical NER dataset, was introduced by \citet{NER_por}, and a Chinese NER dataset was developed by \citet{NER-chinese}. Additionally, NCBI-disease \citep{ner_ncbi} and BC5CDR \citep{ner_BC5CDR} provide annotations for medical entities in PubMed abstracts. To further address limited data, strategies including data augmentation \citep{NER11}, few-shot approaches (\citealp{fewshot}; \citealp{NER14}; \citealp{NER9}), cross-lingual transfer learning (\citealp{NER20}; \citealp{NER4}), and web-based annotation tools \citep{NER_spacy} have been employed.

\section{KBMC : Korean Bio-Medical Corpus}

\subsection{Data Construction}

We use the ChatGPT API\footnote{\url{https://chat.openai.com/}} to create sentences that include medical terminology such as disease names, body parts, and treatments. Given the availability of comprehensive medical domain knowledge and the capabilities of the large language model, we augment the sentences that include medical terminology via responses from gpt-3.5-turbo. The prompts are designed as "Create a Korean sentence comprising more than 20 words that includes \textit{given medical terminology}". All the sentences augmented by ChatGPT undergo thorough review and verification to mitigate the risk of hallucination issues. Medical terms are downloaded from the Korean Standard Terminology Of Medicine (KOSTOM).\footnote{\url{https://www.hins.or.kr/index.es?sid=a1}} It includes 8\textsuperscript{th} revised terms of Korean Standard Classification of Diseases (KCD)\footnote{\url{https://www.koicd.kr/kcd/kcds.do}} and local terms used in the medical field. To facilitate the annotation process, we develop a pre-annotation algorithm that automatically assigns Named Entity tags as a preliminary step.

\begin{figure} 
    \raggedleft     \includegraphics[width=0.5\textwidth]{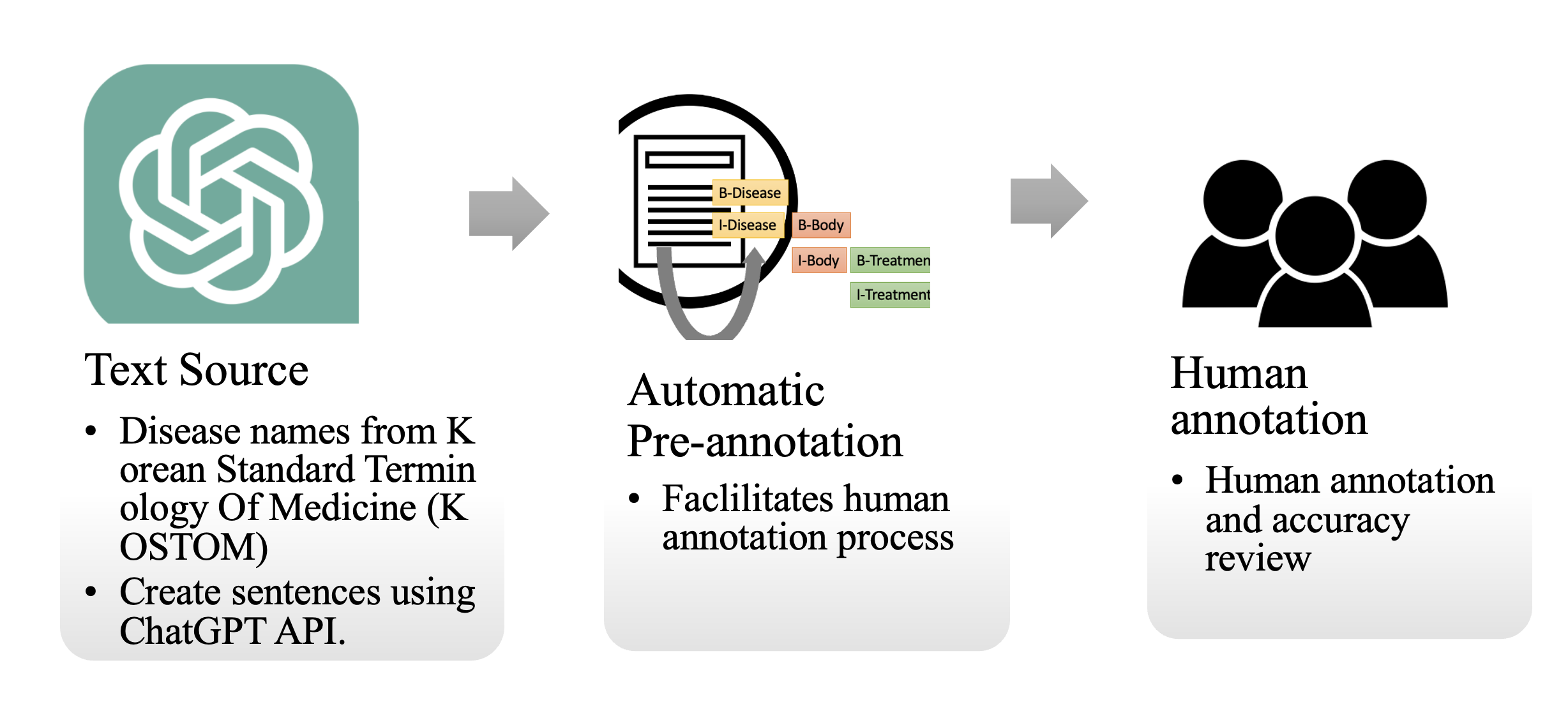}
    \caption{Construction Process of KBMC}
    \label{fig:data_construction}
\end{figure}

\begin{figure} 
    \raggedleft
    \includegraphics[width=0.5\textwidth]{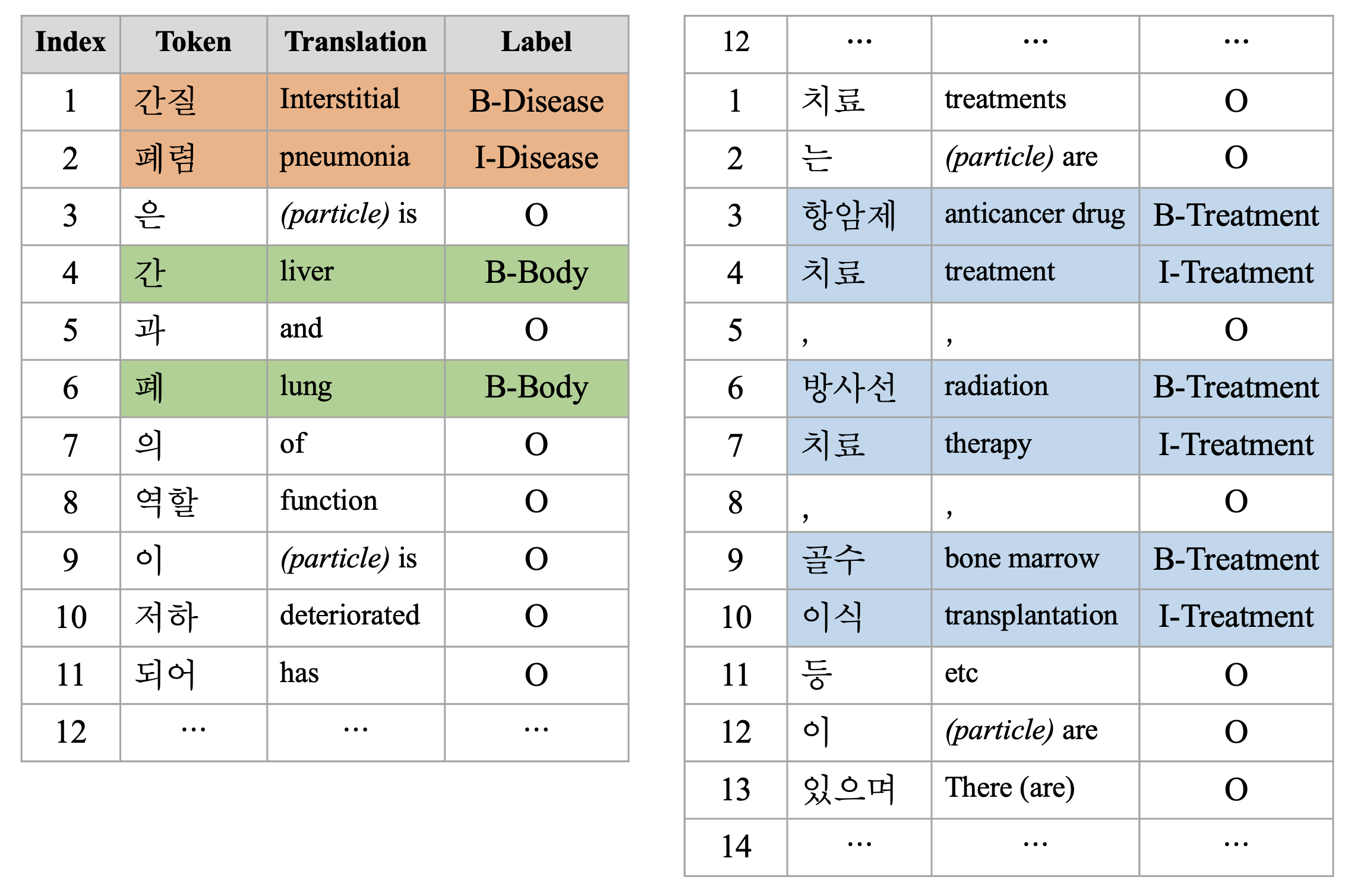}
    \caption{KBMC Annotation}
    \label{fig:KBMC}
\end{figure}

\begin{figure*}
    \centering
    \hspace*{1cm}
    \includegraphics[width=0.8\textwidth]{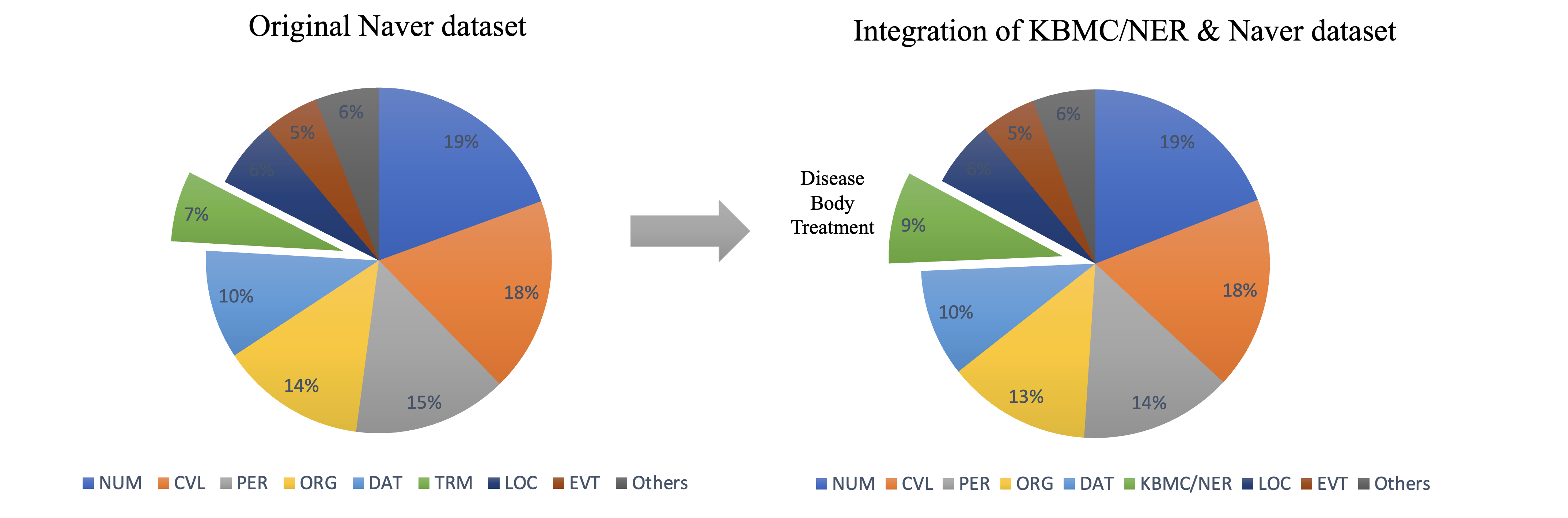}
    \caption{The distribution of Named Entity labels in two datasets: the original Naver NER dataset (left), and a combined version of the Naver NER dataset (partial) and KBMC (right). The original Naver dataset contains the label TRM, representing medical and IT-related terms. In the combined dataset, sentences that include TRM from the original dataset have been replaced with data from KBMC, aiming to achieve a more accurate classification of medical terms into refined categories.}
    \label{fig:piechart}
\end{figure*}

Given a set of collected sentences $W=\{w_{1},w_{2},...,w_{N}\}$, we tokenize each sentence using Open-source Korean Text Processor (OKT)\footnote{\url{https://github.com/open-korean-text/open-korean-text}} so that the input data can be expressed as $\hat{W}=\{x_{1},x_{2},...,x_{M}\}$, where $x_i$ indicates each token. Also, we establish a vocabulary lists for three entity types $E=\{Disease,\ Body,\ Treatment\}$. Then, for each entity type $e\in E$, we detect a set of spans $S_e=\{s_{jk}\mid s_{jk}=\{x_j,...,x_k\}\subset \hat{W}\}$ that matches the vocabulary in the list. Lastly, the algorithm automatically annotates the first token of the span $x_j$ with a $\textbf{B-}$ tag, such as $\textbf{B-Disease}$, and annotates the rest of the tokens with an $\textbf{I-}$ tag, such as $\textbf{I-Disease}$. In other words, $\forall e\in E,\ \forall s_{jk}\in S_{e}$, the pre-annotation of each token $x_i$ can be described as:

\[
\text{Annotate}(x_i) = 
\begin{cases} 
\textbf{B-}e , \text{for } i=j, \\
\textbf{I-}e, \text{for } j<i \leq k
\end{cases}
\]

After the pre-annotation process, four human annotators proceed to annotate Named Entities such as "Disease," "Body," and "Treatment" following the BIO format. The annotators modify and correct any incorrect pre-annotations. Subsequently, a separate fifth reviewer reviews the accuracy of the data annotation for quality control. The annotation process essentially adheres to the standards of the Korean Standard Terminology of Medicine (KOSTOM). However, if there is a disagreement between the annotator and the third annotator (reviewer) regarding the annotation results, the remaining annotators collectively review the mismatched terminology. Figure Figure \ref{fig:data_construction} summarizes the data construction process. Please refer to Appendix \ref{app:A} for example sentences of KBMC.

\subsection{Annotation Result}
KBMC consists of 6,150 sentences, 153,971 tokens in total. Table~\ref{tab:Medical data} displays the label distribution of our dataset. The dataset includes 4,162 distinct disease names, 841 body parts, and 396 treatments. Figure \ref{fig:KBMC} shows the results of the KBMC annotation.

We utilize the OKT (Open-source Korean Text) tokenizer for constructing KBMC. While many Korean NER datasets employ word-level annotation, this approach can be problematic for Korean text. Specifically, word-level tokenization often fails to distinguish between nouns and associated postpositional particles, leading to imprecise annotations by attributing a single Named Entity tag to combined terms and particles. Given that Korean is an agglutinative language, tokenizing at the morpheme level is more precise. Thus, unlike conventional Korean NER datasets, we tokenize sentences into morphemes to ensure more accurate annotations.
\begin{table}
    \centering
    \begin{tabular}{l|l|c}
         \Xhline{3\arrayrulewidth}
         \textbf{Named Entity (NE)} & \textbf{Scheme} & \textbf{\# of NE}\\
         \hline
         \multirow{2}{*}{\textbf{Disease}} & B (Begin) & 10,595\\
         & I (Inside) & 10,089\\
         \hline
         \multirow{2}{*}{\textbf{Body}} & B (Begin) & 5,215\\
         & I (Inside) & 1,158\\
         \hline
         \multirow{2}{*}{\textbf{Treatment}} & B (Begin) & 1,193\\
         & I (Inside) & 839\\
         \Xhline{3\arrayrulewidth}
    \end{tabular}
    \caption{Label Distribution of KBMC }
    \label{tab:Medical data}
\end{table}
\label{sec:KBMC}

\subsection{Data Application}
For data augmentation and comparison of NER in general and domain-specific text, the Naver NER dataset\footnote{\url{https://github.com/naver/nlp-challenge/tree/master/missions/ner}} is concatenated with KBMC. The Naver NER dataset is a general NER dataset, published by Naver\footnote{\url{https://www.navercorp.com/}} and Changwon University.
The Naver NER dataset comprises 90,000 sentences and includes 14 named entities, such as PER (Person), FLD (Field), NUM (Number), DAT (Date), and ORG (Organization). Specifically, the dataset includes annotated named entities labeled as TRM (TERM), which refer to medicine and IT-related terminology. To prevent any potential mismatches when concatenated with our KBMC, we exclude 12,426 sentences containing TRM from the Naver NER dataset. The concatenated version of the Naver NER dataset and KBMC includes 13 general Named Entities and 3 medical Named Entities, totaling 16 Named Entities. The integration of the datasets and their label distribution are demonstrated in Figure \ref{fig:piechart}.

\section{KBMC}
In this section, we compare the performance between the utilization of the Naver dataset, also known as a general NER dataset, and the application of KBMC. Next, to assess the applicability of KBMC, we conduct NER using MedSpaCy\footnote{\url{https://github.com/medspacy/medspacy}}.

\subsection{Models}
The data variation experiment confirms and quantifies the impact of the training data. We evaluate our dataset on six different language models: KM-BERT \citep{kmbert}, KR-BERT \citep{krbert}, KoBERT\footnote{\url{https://github.com/SKTBrain/KoBERT}}, KR-ELECTRA\footnote{\url{https://github.com/snunlp/KR-ELECTRA}}, KoELECTRA v3\footnote{\url{https://github.com/monologg/KoELECTRA}}, and BiLSTM-CRF \citep{BiLSTM-CRF}. These are advanced Korean NLP models, each with unique architectures and approaches to understanding language. While KR-BERT, KoBERT, KR-ELECTRA, and KoELECTRA v3 leverage transformer-based architectures to achieve state-of-the-art performance on various NLP tasks, BiLSTM-CRF combines bidirectional long short-term memory units with a conditional random field layer, catering to tasks such as NER. Among these models, KM-BERT is a domain-specific language model that has been trained on the Korean medical corpus.
Both the learning rate (ranging from 1e-5 to 5e-5) and the batch size (ranging from 32 to 128) are adjusted for optimal performance.

\begin{table}
\centering
\resizebox{0.48\textwidth}{!}
{%
\begin{tabular}{l|c|c|c}
\Xhline{3\arrayrulewidth}
\textbf{Model}       & \textbf{Avg.F1(General)} & \textbf{medical NE} & \textbf{F1 of medical NER} \\
\hline
KM-BERT & 87.08         & TRM         & 75.35           \\
{\citep{kmbert}} & {} & {} & {}\\
\hline
{KR-BERT} & \multirow{2}{*}{86.51} & \multirow{2}{*}{TRM} & \multirow{2}{*}{75.26}\\
{\citep{KR-BERT}} & {} & {} & {}\\
\hline
{Ko-BERT} & {88.01} & {TRM} & {78.21}\\
\hline
KR-ELECTRA  & \multirow{2}{*}{87.62} & \multirow{2}{*}{TRM} & \multirow{2}{*}{76.25}\\
{\citep{kr-electra}} & {} & {} & {}\\
\hline
Ko-ELECTRA  & 88.00         & TRM         & 76.58             \\
\hline
BiLSTM-CRF  & \multirow{2}{*}{55.23} & \multirow{2}{*}{TRM} & \multirow{2}{*}{42.23}\\
{\citep{BiLSTM-CRF}} & {} & {} & {}\\
\Xhline{3\arrayrulewidth} \end{tabular}}
\caption{Medical Named Entities and NER Performance: General NER dataset (The Naver Dataset) solely used.}
\label{tab:experiment2_Naver}
\end{table}

\subsection{Results}
\subsubsection*{Medical NER using general dataset}

We initially fine-tune six language models using the Naver dataset, which primarily contains general labels. For the experiments, the dataset is split into 90\% for training and 10\% for testing. All medical entities in this dataset are grouped under one label, TRM. However, this label is not solely for medical terms; it also includes IT-related entities. This generalization makes it difficult to accurately identify and differentiate medical terms since they are consolidated with IT terms under TRM. As a result, the identification of specific medical terminology becomes challenging. Additionally, the F1 score for medical NER using the Naver dataset is below average, as indicated in Table~\ref{tab:experiment2_Naver}.

\begin{table}
\centering
\resizebox{0.5\textwidth}{!}
{%
\begin{tabular}{l|c|c|c}
\Xhline{3\arrayrulewidth}
\textbf{Model}       & \textbf{Avg.F1(General)} & \textbf{Medical NEs} & \textbf{F1 of Medical NER} \\
\hline
\multirow{3}{*}{KM-BERT} & \multirow{3}{*}{88.53 \textcolor{blue}{(+1.45)}} & {Disease} & {98.04 \textcolor{blue}{(+22.69)}}\\
        & {} & {Body} & {98.13 \textcolor{blue}{(+22.78)}}\\
        & {} & {Treatment} & {98.53 \textcolor{blue}{(+23.18)}}\\
\hline
\multirow{3}{*}{KR-BERT} & \multirow{3}{*}{87.48 \textcolor{blue}{(+0.97)}} & {Disease} & {98.04 \textcolor{blue}{(+22.78)}}\\
        & {} & {Body} & {98.32 \textcolor{blue}{(+23.06)}}\\
        & {} & {Treatment} & {97.82 \textcolor{blue}{(+22.56)}}\\
\hline
\multirow{3}{*}{KoBERT} & \multirow{3}{*}{88.70 \textcolor{blue}{(+0.69)}} & {Disease} & {98.25 \textcolor{blue}{(+20.04)}}\\
        & {} & {Body} & {98.22 \textcolor{blue}{(+20.01)}}\\
        & {} & {Treatment} & {98.18 \textcolor{blue}{(+19.97)}}\\
\hline
\multirow{3}{*}{KR-ELECTRA} & \multirow{3}{*}{88.63 \textcolor{blue}{(+1.01)}} & {Disease} & {98.21 \textcolor{blue}{(+21.96)}}\\
        & {} & {Body} & {98.31 \textcolor{blue}{(+22.06)}}\\
        & {} & {Treatment} & {98.53 \textcolor{blue}{(+22.28)}}\\
\hline
\multirow{3}{*}{KoELECTRA} & \multirow{3}{*}{88.86 \textcolor{blue}{(+0.86)}} & {Disease} & {98.05 \textcolor{blue}{(+21.47)}}\\
        & {} & {Body} & {97.72 \textcolor{blue}{(+21.14)}}\\
        & {} & {Treatment} & {96.56 \textcolor{blue}{(+19.98)}}\\
\hline
\multirow{3}{*}{BiLSTM-CRF} & \multirow{3}{*}{56.68 \textcolor{blue}{(+1.45)}} & {Disease} & {88.18 \textcolor{blue}{(+45.95)}}\\
        & {} & {Body} & {81.44 \textcolor{blue}{(+39.21)}}\\
        & {} & {Treatment} & {61.14 \textcolor{blue}{(+18.91)}}\\
\Xhline{3\arrayrulewidth}
\end{tabular}}
\caption{Medical Named Entities and Performance:  KBMC applied. The numbers in blue indicate the degree of improvement when compared to the experimental results in Table~\ref{tab:experiment2_Naver}.}
\label{tab:experiment2_KBMC}
\end{table}

% \begin{table}
%     \centering
%     \begin{tabular}{{c|c|c|c}}
%     \Xhline{3\arrayrulewidth}
% \textbf{Model}       & \textbf{Avg. F1} & \textbf{Medical NEs} & \textbf{F1 of Medical NER} \\
% \hline
% \multirow{3}{*}{KM-BERT} & \multirow{3}{*}{88.53} & {Disease} & {98.04}\\
%         & {} & {Body} & {98.13}\\
%         & {} & {Treatment} & {98.53} \\
%          \Xhline{3\arrayrulewidth}
%     \end{tabular}
%     \caption{Performance of KM-BERT \citep{kmbert} using KBMC both (medical) and Naver NER dataset (general)}
%     \label{KM-BERT}
% \end{table}
% \label{KM-BERT}

\subsubsection*{Medical NER using KBMC}
We introduce KBMC to address the shortcomings of the Naver dataset. By combining the Naver NER dataset (excluding sentences with TRM) with KBMC, we achieve a more balanced dataset. The average F1 score in Table~\ref{tab:experiment2_KBMC} encompasses 13 general entities from the Naver dataset (TRM excluded) and 3 medical entities from KBMC. The dataset is divided into 90\% for training and 10\% for testing. To avoid data imbalance, we maintain consistent proportions of general and medical data in both training and testing phases. KBMC offers precise categorization, separating medical entities from IT-related ones and allowing for detailed classification. This specificity results in a performance increase, with the F1 scores for Disease, Body, and Treatment labels surpassing the TRM label by nearly 20 points. The consistent performance across different models demonstrates the quality and reliability of our dataset.

\begin{table}
    \centering
    \begin{tabular}{l|ccc}
         \Xhline{3\arrayrulewidth}
         {}&{\textbf{Avg.F1}} & \textbf{Precision} & \textbf{Recall}\\
         \hline
         \textbf{MedSpaCy}&{95.69} & {97.02} &{95.52}\\
         \Xhline{3\arrayrulewidth}
    \end{tabular}
    \caption{Performance of MedSpaCy NER using KBMC}
    \label{MedSpaCy}
\end{table}
\label{sec:MedSpaCy}

\subsection{KBMC Applicability Assessment}

\subsubsection*{Medical NER using MedSpaCy}
In order to test the utility of KBMC, we also test our dataset using MedSpaCy. \citet{medspacy} have released a library of tools for clinical NLP and text processing with SpaCy.\footnote{\url{https://spacy.io/}} We apply MedSpaCy on Korean dataset by using ko\_core\_news\_md,\footnote{\url{https://spacy.io/models/ko\#ko_core_news_md}} a pretrained statistical model for Korean provided by SpaCy. As shown in Table \ref{MedSpaCy}, our KBMC dataset demonstrates remarkable performance on a clinical text processing toolkit in Python as well. While MedSpaCy may not be primed for general entity recognition, it excels in identifying medical terms, especially when enhanced with KBMC.

\section{Conclusion}
In our research, we introduce KBMC, the first open-source biomedical NER dataset tailored for the Korean language. KBMC provides a training ground for language models to detect and categorize medical Named Entities, addressing the issue of data scarcity in this domain.

We evaluate the utility of the KBMC dataset in two scenarios: one using only a pre-existing general NER dataset, and another incorporating the KBMC dataset. The inclusion of KBMC resulted in enhanced predictions for medical Named Entities and an elevated overall F1 score, which averages the F1 scores for both general and medical entities. With KBMC, models can recognize a broader spectrum of medical terms. Notably, when paired with MedSpaCy, a Python toolkit designed for clinical NLP, our dataset showcases impressive results.

We anticipate that our KBMC dataset will contribute substantially to ongoing research in the field of medical NLP.

\section*{Limitations}
The primary challenge arises from the limited availability of Korean medical data, which makes it difficult to develop a comprehensive corpus. Due to this constraint, we were unable to manually create a labeled dataset for downstream tasks other than NER task. As a result, an important avenue for future research lies in the construction of a more expansive and diverse Korean medical corpus to facilitate the development of other downstream tasks, such as question-answering (QA). Moreover, while our intention was to compare different general NER datasets in terms of medical entity extraction, The Naver dataset was the only available Korean NER dataset that provided annotations for medical terminology. This kind of problem also occurred in terms of domain-specific models as well. KM-BERT was the only medical language model available for our testing. This limited access to resources restricted our capacity for a comprehensive comparison.

\section*{Ethics Statement}
Using our KBMC dataset enables precise identification of entity categories. When implemented in the medical sphere, our dataset and model can assist in de-identifying personal details of patients. In the realm of medical NLP, transferring and accessing data is challenging due to the presence of sensitive content. To address these privacy and data sensitivity issues, integrating medical NER into real-world medical institutions offers a safeguarded approach. Resolving these challenges sets the stage for a flourishing future in NLP research, spanning areas such as the medical and legal fields.

\bibliographystyle{lrec-coling2024-natbib.bst}
\bibliography{lrec-coling2024-example}

\newpage
\appendix
\section*{Appendix}
\section{Appendix}\label{app:A}

\begin{table}[h]
\centering
\begin{tabularx}{\textwidth}{|X|X|X|}
\Xhline{3\arrayrulewidth}
\textbf{KBMC sentences} & \textbf{Translation} & \textbf{NER Tags} \\
\Xhline{3\arrayrulewidth}
\begin{CJK}{UTF8}{mj} 전신 적 다한증 은 신체 전체 에 힘 이 빠져서 일상 생활 이 어려워지는 질환 으로 , 근육 통증 과 무기 력 감 이 동반 됩니다 .\end{CJK} & Systemic myasthenia is a condition in which the whole body loses strength, making daily life difficult, accompanied by muscle pain and a sense of lethargy. & Disease-B Disease-I Disease-I O O O O O O O O O O O O O O Disease-B Disease-I O Disease-B Disease-I Disease-I O O O O \\
\hline
\begin{CJK}{UTF8}{mj}췌장암 이란 췌장 에 생긴 암세포 로 이루어진 종괴 ( 종양 덩어리 ) 이다 . \end{CJK}& Pancreatic cancer refers to a tumor (a lump of tumor) made up of cancer cells that form in the pancreas. & Disease-B O Body-B O O O O O Disease-B O Disease-B O O O O \\
\hline
\begin{CJK}{UTF8}{mj}이러한 병명 은 폐 기능 저하 로 인한 호흡 곤란 기침 천식 발작 등 의 증상 을 유발 하 여 일상생활 에 큰 영향 을 미칩니다 .\end{CJK} & Such diseases lead to symptoms such as respiratory distress, coughing, asthma attacks, etc., caused by decreased lung function, greatly affecting daily life. & O O O Disease-B Disease-I Disease-I Disease-I Disease-I Disease-I Disease-I Disease-I Disease-I Disease-I O O O O O O O O O O O O O O \\
\hline
\begin{CJK}{UTF8}{mj}버킷 림프종 은 림프절 에서 발생 하는 악성 종양 으로 , 조기 발견 과 치료 가 중요하며 항암 치료 나 방사선 치료 등 다양한 치료법 이 존재 합니다 .\end{CJK} & Burkitt lymphoma is a malignant tumor that originates in the lymph nodes. Early detection and treatment are crucial, and various treatment methods, such as chemotherapy and radiation therapy, exist.&	Disease-B Disease-I O Body-B O O O Disease-B Disease-I O O O O O O O O Treatment-B Treatment-I O Treatment-B Treatment-I O O O O O O O\\

\Xhline{3\arrayrulewidth}
\end{tabularx}
\caption{Examples of the KBMC dataset}
\end{table}

\end{document}